\documentclass[10pt,journal,compsoc]{IEEEtran}

\usepackage{amsmath}
\usepackage{amssymb}
\usepackage{epsfig}
\usepackage{graphicx}
\usepackage{xcolor}
\usepackage{tablefootnote}

\newtheorem{thm}{Theorem}
\newtheorem{lemma}[thm]{Lemma}

\ifCLASSOPTIONcompsoc
  \usepackage[nocompress]{cite}
\else
  \usepackage{cite}
\fi
\ifCLASSINFOpdf
\else
\fi

\begin{document}
%
\title{Integrating Multiple Receptive Fields\\ through Grouped Active Convolution}
%
%
%
%

\author{Yunho~Jeon 
        and~Junmo~Kim,~\IEEEmembership{Member,~IEEE}
\IEEEcompsocitemizethanks{\IEEEcompsocthanksitem Y. Jeon and J. Kim are with the School of Electrical Engineering, Korea Advanced Institute of Science and Technology, Daejeon, Korea.\protect\\
E-mail: jyh2986@kaist.ac.kr, junmo.kim@kaist.ac.kr}
\thanks{Manuscript received XXXX XX, XXXX; revised  XXXX XX, XXXX}}

%
%

\markboth{Journal of \LaTeX\ Class Files,~Vol.~14, No.~8, August~2015}%
{Jeon \MakeLowercase{\textit{et al.}}: Integrating Multiple Receptive Fields\\ through Grouped Active Convolution}
%



\IEEEtitleabstractindextext{%
	\begin{abstract}
		Convolutional networks have achieved great success in various vision tasks. This is mainly due to a considerable amount of research on network structure. In this study, instead of focusing on architectures, we focused on the convolution unit itself. The existing convolution unit has a fixed shape and is limited to observing restricted receptive fields. In earlier work, we proposed the active convolution unit (ACU), which can freely define its shape and learn by itself. In this paper, we provide a detailed analysis of the previously proposed unit and show that it is an efficient representation of a sparse weight convolution. Furthermore, we extend an ACU to a grouped ACU, which can observe multiple receptive fields in one layer. We found that the performance of a naive grouped convolution is degraded by increasing the number of groups; however, the proposed unit retains the accuracy even though the number of parameters decreases. Based on this result, we suggest a depthwise ACU, and various experiments have shown that our unit is efficient and can replace the existing convolutions.
	\end{abstract}
	
	\begin{IEEEkeywords}
		Convolutional neural network (CNN), Multiple Receptive Fields, depthwise convolution, deep learning.
\end{IEEEkeywords}}

\maketitle

\IEEEdisplaynontitleabstractindextext

%
\IEEEpeerreviewmaketitle

\ifCLASSOPTIONcompsoc
\IEEEraisesectionheading{\section{Introduction}\label{sec:introduction}}
\else
\section{Introduction}
\label{sec:introduction}
\fi

%
%
%
%
\IEEEPARstart{C}{onvolutional} neural network (CNN) has become a major topic of deep learning, especially in visual recognition tasks. After the great success at the ImageNet Large Scale Visual Recognition Challenge (ILSVRC) of 2012\cite{ILSVRC15}, many efforts have been made to improve accuracy while reducing computational budgets by using CNN. The major focus for this research was on designing network architectures~\cite{Simonyan2015,szegedy2017inception,szegedy2015going,szegedy2016rethinking,he2016deep,he2016identity}. Recently, attempts were made to automatically generate efficient network architectures \cite{zoph2016neural, zoph2017learning}, and the generated networks achieved a better result than the conventional networks. This approach is yet very slow and difficult to train by using feasible amounts of resources but will affect the designing of networks. In such studies, components can be considered as more important factors than network construction.

Some other studies have focused on components to improve the performance of the network. Such methods suggest replacing existing units with new components while retaining the network architectures. In the early years of the development of this method, researchers focused mainly on activation units~\cite{nair2010rectified, maas2013rectifier, xu2015empirical, he2015delving, ClevertUH15}. Variants of pooling were also proposed in many studies~\cite{graham2014fractional, lee2016generalizing, he2014spatial, girshick2015fast}. Although the convolution unit is a core component of CNN, research on this unit has only been under way in recent years. Many variants of convolutions have been proposed to overcome weaknesses of the conventional convolution\cite{worrall2017harmonic, zhou2017oriented,chen2014semantic,YuKoltun2016, jia2016dynamic, dai2017deformable}. 

As the shape of a convolution unit is fixed, it can only observe restricted receptive fields. Chen et al. \cite{chen2014semantic} and Yu and Koltun \cite{YuKoltun2016} suggested a dilated convolution, which can expand the receptive field by retaining the parameters; this was applied for the dense prediction of segmentation. However, its dilation was fixed and set on initialization. In our previous paper~\cite{jeon2017active}, we solved this problem by introducing position parameters and called the unit as an active convolution unit (ACU). The proposed unit could define any shape of convolution, and the shape is learnable and can be optimized through backpropagation.

\begin{figure}
	\centering	
	\begin{tabular}{p{0.25\linewidth}p{0.25\linewidth}p{0.3\linewidth}}
		\multicolumn{3}{c}{\includegraphics[width=0.85\linewidth]{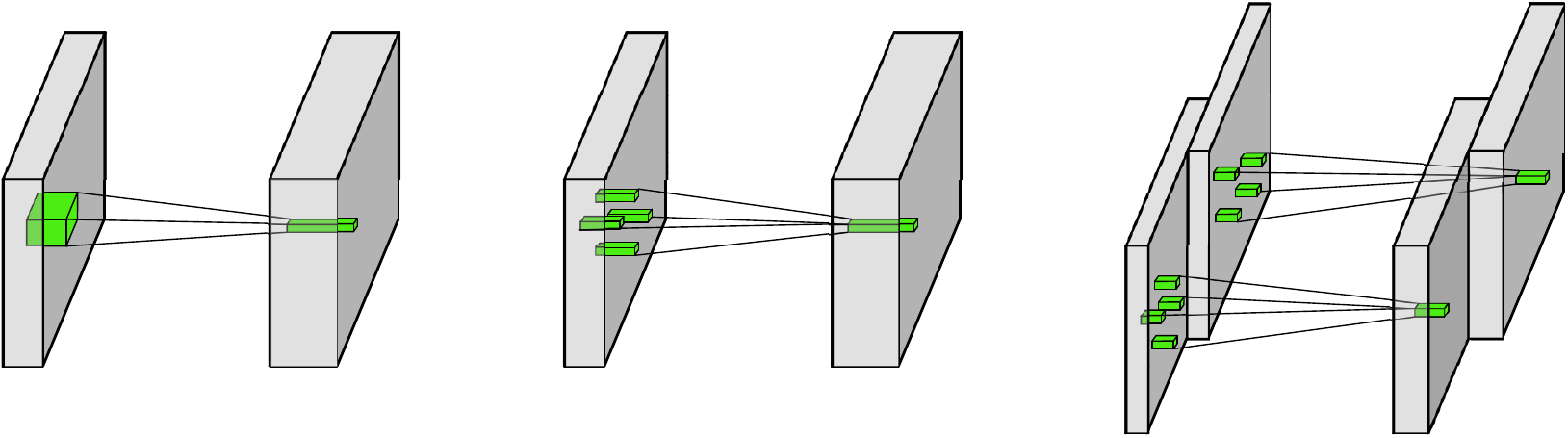}}\smallskip\\
		\centering\shortstack{(a) Naive\\ convolution} & 
		\centering\shortstack{(b) ACU \\ \phantom{}} & 
		\centering\shortstack{(c) Grouped\\ ACU}
	\end{tabular}	
	\caption{Comparison of different types of convolution units. (a) Conventional convolution unit has a fixed shape. (b) In comparison, ACU can learn its shape, which is shared for all outputs. (c) Grouped ACU can have multiple shapes and view different receptive fields according to its groups.}\label{fig:ACU_concept}
\end{figure}

Attempts have been made to view multiple receptive fields by configuring several units in parallel for improving accuracy. Inception\cite{szegedy2017inception,szegedy2015going,szegedy2016rethinking} is a module using different sizes of convolutions. Pyramid pooling\cite{he2014spatial} and atrous pooling\cite{chen2018deeplab,chen2017rethinking} have also been suggested to aggregate features at different scales. Although the ACU could optimize its shape, it could possess only one shape for each layer because that shape is shared for all output channels. To overcome this weakness, we developed a grouped ACU inspired from grouped convolution\cite{xie2017aggregated,zhang2017shufflenet} and depthwise convolution\cite{chollet2016xception,howard2017mobilenets, sandler2018inverted}. A grouped ACU can have more than one shape and integrate multiple receptive fields in one layer. To the best of our knowledge, this is the first component that is able to view multiple different receptive fields in one layer without concatenating outputs from multiple components.


In this paper, we further consolidate the results of a previous research\cite{jeon2017active} through further analysis and experiments. Moreover, we extended the ACU to a grouped ACU, which can view multiple receptive fields in one layer. Through this study, we show that the proposed unit is an efficient component for the formation of network topologies. The key contributions of this paper are summarized as follows:
\begin{itemize}
	\item In this study, we performed additional analysis on ACU and showed that the ACU is an efficient representation of a sparse weight convolution. Further experiments explain why ACU can achieve better results with similar numbers of parameters.
	
	\item 
	The grouped ACU is proposed, which is the combination of the original ACU and grouped convolution. While the original ACU uses only one shape and shares it for all channels, the grouped ACU uses multiple position sets in a layer. It is able to receive multiple receptive fields at one layer while reducing the computational complexity. Moreover, we show that the Inception module can be simplified through this unit. 
	
	\item
	With the application of our proposed ACUs in various network architectures, we show that the units are simple, efficient, and can replace the existing convolutions.
	
\end{itemize}

In the next section, we review the related works on network architectures and the variants of convolution units. In Section~\ref{sec:revise ACU}, we revisit ACU, which was proposed in \cite{jeon2017active}, and provide a detailed analysis of the characteristics of the unit. In Section~\ref{sec:Grouped ACU}, we propose grouped ACU, which divides the input and output channels according to a group and shares position parameters in each group, and extend the proposed unit to depthwise ACU. In Section~\ref{sec:experiment}, experimental results are demonstrated to show the effectiveness of the proposed method.

\section{Related Work}
Our approach is based on the success of CNN for image classifications. The methodology of such a classification has spread to various other applications including semantic segmentation~\cite{chen2018deeplab,long2015fully,chen2014semantic} and object detection~\cite{girshick2014rich,ren2015faster,redmon2016you,liu2016ssd}.

\medskip \noindent
\textbf{CNN Architectures}
Most research on CNN has focused on developing an architecture to achieve a better result. AlexNet~\cite{Krizhevsky2012} uses various types of convolutions and stacks some pooling layers to reduce spatial dimensions. VGG~\cite{Simonyan2015} is based on the idea that a stack of two $3\times3$ convolution layers is more effective than $5\times5$ layers. This network is used broadly for many applications owing to the simplicity of the topology. GoogleNet~\cite{szegedy2017inception,szegedy2015going,szegedy2016rethinking} introduced an Inception layer for the composition of various receptive fields. This network showed that a carefully crafted design can achieve a better result while maintaining a similar computational budget. The residual network~\cite{he2016deep,he2016identity,zagoruyko2016wide} enables training of overly deep networks by adding shortcut connections to implement identity mapping, allowing for deeper networks to be configured. Later, many variants of a residual network were proposed\cite{han2017deep,chen2017dual,hu2017squeeze}.

\medskip \noindent
\textbf{Variants of Convolution}
The dilated convolution~\cite{chen2014semantic,YuKoltun2016} was suggested to enhance the resolution of the result and reduce postprocessing in semantic segmentation tasks. In our previous work~\cite{jeon2017active}, we introduced the ACU, which is a generalization of the naive convolution. By introducing position parameters, any shape of convolution can be defined, and the shape can be learned through formal backpropagation.

Some studies have attempted to use dynamic weights instead of fixed weights. Dynamic filter network~\cite{jia2016dynamic} convolves the weight received from other networks. Deformable convolution~\cite{dai2017deformable} is another type of dynamic network, which receives the position parameter from other networks, and the convolution shape is changed dynamically. The concept of deformable convolution is similar to our ACU. The major difference is that ACU possesses intrinsic position parameters, while a deformable convolution receives the position parameters from other networks, and thus needs another network to generate them. 

\medskip \noindent
\textbf{Combination of Multiscale Features}
Beyond extending the areas of receptive fields, there have been many attempts to combine multiple fields. Spatial pyramid pooling~\cite{he2014spatial} was suggested for integrating receptive fields at different scales. GoogleNet~\cite{szegedy2017inception,szegedy2015going,szegedy2016rethinking} formed the Inception module composed of multiple size convolutions; this can create better features by using less number of parameters. In the segmentation tasks, Deeplab~\cite{chen2018deeplab,chen2017rethinking} used the atrous spatial-pyramid-pooling layer, which uses multiple filters with different dilation rates. All of these works comprised multiple operations, and no unified component was built that allows the viewing of multiple receptive fields at different scales.

\medskip \noindent
\textbf{Grouped Convolution}
Grouped convolution divides its input and output into small groups, and calculates the outputs by using only input channels within the same group. AlexNet~\cite{Krizhevsky2012} first introduced the grouped convolution not for improving accuracy but for other practical reasons, i.e., for running a large network in restricted resources, convolution was divided into two groups. In ShuffleNet~\cite{zhang2017shufflenet}, a grouped convolution is used for reducing the computational cost while maintaining accuracy with limited resources on the embedded devices. ResNeXt~\cite{xie2017aggregated} shows that increasing the number of groups is more efficient than expanding depth or width for gaining accuracy. 

\medskip \noindent
\textbf{Depthwise Convolution}
Depthwise (or channelwise) convolution is a special case of grouped convolution, in which the number of input and output channels is the same as the number of groups. This implies that each output channel is calculated using only a corresponding input channel. Xception~\cite{chollet2016xception} uses depthwise separable convolution, which first applies the depthwise convolution, followed by the pointwise convolution. This operation reduces the number of weight parameters efficiently, and the corresponding network achieves a better result with fewer parameters. MobileNet~\cite{howard2017mobilenets, sandler2018inverted} also employs the depthwise convolution to reduce the network size and run fast in embedded devices.

\begin{figure*}
	\centering	
	\begin{tabular}{p{0.45\textwidth}p{0.45\textwidth}}
		\multicolumn{2}{c}{\includegraphics[width=0.9\textwidth]{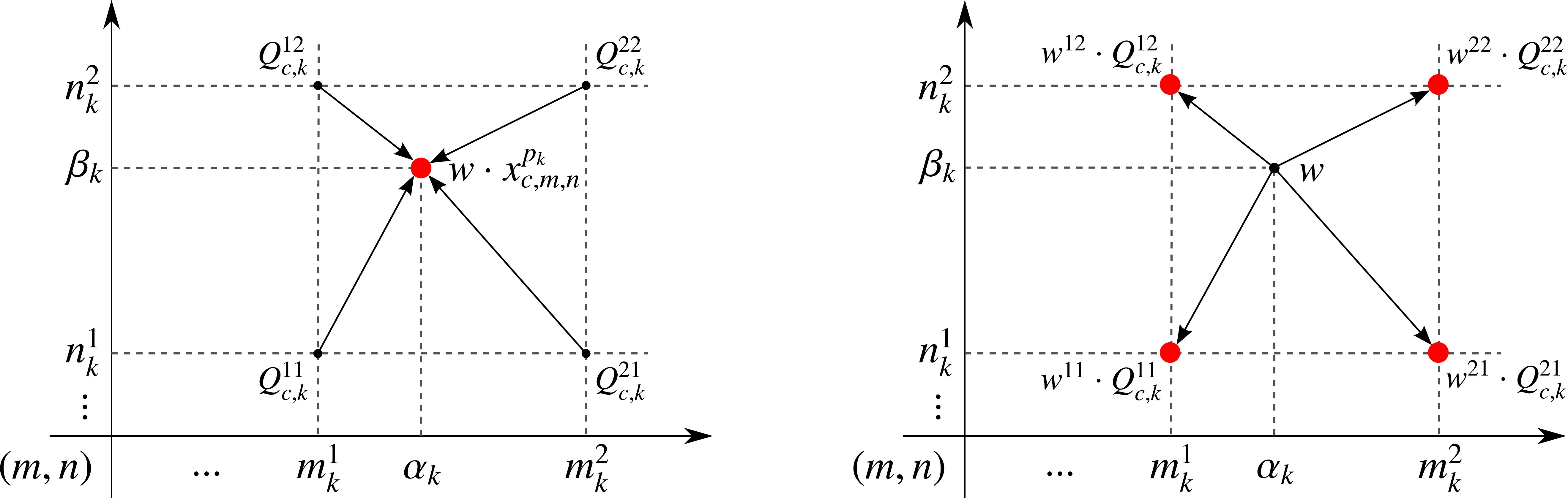}}\smallskip\\
		\centering (a) Calculation using interpolated input& \centering (b) Calculation using extrapolated weights\\
	\end{tabular}	
	\caption{Comparison of two methods for calculating ACU. (a) Weights are multiplied with interpolated inputs. (b) Extrapolated weights are multiplied with original inputs, and summed up. These two methods obtain the same result.}
	\label{fig:calc_by_interpolation_extrapolation}
\end{figure*}

\section{Active Convolution}\label{sec:revise ACU}
In this section, we will revisit the basic concept of the ACU proposed in an earlier work~\cite{jeon2017active}. Naive convolution unit has learnable weights $\boldsymbol{W}$, and each weight is convolved with its input (Eq.~\eqref{eq:conv eq1}). This can be written for one output value $y_{m,n}$ in given location $(m,n)$, as shown in Eq.~\eqref{eq:conv eq2}. We omitted a bias term and assumed one output channel for simplicity. 

\begin{equation}
\label{eq:conv eq1}
\boldsymbol{Y} = \boldsymbol{W} * \boldsymbol{X}
\end{equation}
\begin{equation}
\label{eq:conv eq2}
y_{m,n} = \sum_{c}\sum_{i,j} w_{c,i,j} \cdot x_{c,m+i,n+j}
\end{equation}
where $c$ is the index of the input channel, and $w_{c,i,j}$ and ${x}_{c,m,n}$ are the weight and input value in the given channel $c$ and position $(m,n)$, respectively. Index $i,j$ defines the fixed convolving area (e.g. $i,j \in \{-1,0,1\}$ for 3$\times$3 convolution).

Unlike naive convolution, ACU has the additional position parameter $\theta_p$, which defines the horizontal and vertical displacement ($\alpha_k$, $\beta_k$) of the input from the center of the filter (Eq.~\eqref{eq:position param}). We denote one acceptor of the ACU as \textit{synapse}, and each synapse has its own weight and position.
\begin{equation}
\begin{aligned}
\label{eq:position param}
\theta_p = \{p_k|0\leq k < K\},\\
p_k = (\alpha_k, \beta_k) \in \mathbb{R}^2
\end{aligned}
\end{equation}
where $k$ is the index of the synapse. By using this position parameter, the ACU is defined as

\begin{equation}
\label{eq:ACU eq1}
\boldsymbol{Y} = \boldsymbol{W} * \boldsymbol{X}_{\theta_p}
\end{equation}
\begin{equation}
\label{eq:ACU eq2}
\begin{aligned}
y_{m,n} = &\sum_{c}\sum_{k} w_{c,k} \cdot {x}^{p_k}_{c,m,n}\\
= &\sum_{c}\sum_{k} w_{c,k} \cdot x_{c,m+\alpha_k,n+\beta_k}
\end{aligned}
\end{equation}
where ${x}^{p_k}_{c,m,n}$ is the input value located at displacement $p_k$ from position $(m,n)$ at channel $c$. 

$\alpha_k$ and $\beta_k$ are not limited to integers but are allowed to be real numbers. When these values are real, the location of input is placed in an inter-lattice point. The value of this point can be calculated through interpolation. By using bilinear interpolation, the output value is differentiable according to $\alpha_k$ and $\beta_k$, and these parameters are learnable through backpropagation. Backward calculation for the weight and bias is the same as that for the naive convolution, and the derivatives for the position parameter can be calculated easily. Please refer to \cite{jeon2017active} for more details.

\subsection{What is an ACU?}

In a previous study \cite{jeon2017active}, we defined a new form of convolution by adding position parameters and showed that this new unit improves classification accuracy. Thus, the following questions must be answered: is ACU a new operation that is completely different from convolution? What characteristics of the ACU improve the performance of the network? In this section, we answer these questions and show that ACU is actually the efficient representation of a sparse weight convolution.

\smallskip
\begin{lemma}
	\label{thm:conv is subset of acu}
	The set containing all operations represented by convolution is the subset of the set containing all operations represented by ACU, i.e., $\boldsymbol{C} \subset \boldsymbol{A}$
\end{lemma}
where $\boldsymbol{A}$ is the set containing all operations represented by ACU, and $\boldsymbol{C}$ is the set containing all operations represented by convolution. Clearly, any convolution can be represented by an ACU. Given a convolution, we can convert it to an ACU by using the same weight, and assign position parameters based on the given shape. For instance, the conventional $3\times3$ convolution is represented by the ACU with $\theta_p = \{(-1,-1), (0,-1), $ $(1,-1),(-1,0),(0,0),(1,0),(-1,1),(0,1),(1,1)\}$.

\smallskip
\begin{thm}
	\label{thm:equivalent calculation for ACU}
	Given position parameter $\theta_p$ and weight $\boldsymbol{W}$, there exists an extrapolated weight $\boldsymbol{\overline W}_{\theta_p}$ which holds for
	
	\centerline{$\boldsymbol{W} * \boldsymbol{X}_{\theta_p} = \boldsymbol{\overline W}_{\theta_p} * \boldsymbol{X}$}
\end{thm}
The detailed proof of Theorem~\ref{thm:equivalent calculation for ACU} is in Appendix~\ref{APX:mathmatcial expansion}, and here we provide conceptual explanations. We first consider an ACU with one synapse. Fig.~\ref{fig:calc_by_interpolation_extrapolation}(a) shows the original calculation using interpolating inputs. Interpolated input ${x}^{p_k}_{c,m,n}$ is derived using four neighbor input values, and weight $w$ of the given synapse is multiplied with the interpolated input. In Fig.~\ref{fig:calc_by_interpolation_extrapolation}(b), instead of interpolating inputs, the weight is extrapolated to the nearest four neighbors. Each extrapolated weight $w^{ab}$ is multiplied with the original input value, and the summation of these four values provides the same result as that obtained through the previous calculation. This procedure in Fig.~\ref{fig:calc_by_interpolation_extrapolation}(b) can be regarded as a 2$\times$2 convolution.

\begin{figure}
	\centering
	\includegraphics[width=0.45\textwidth]{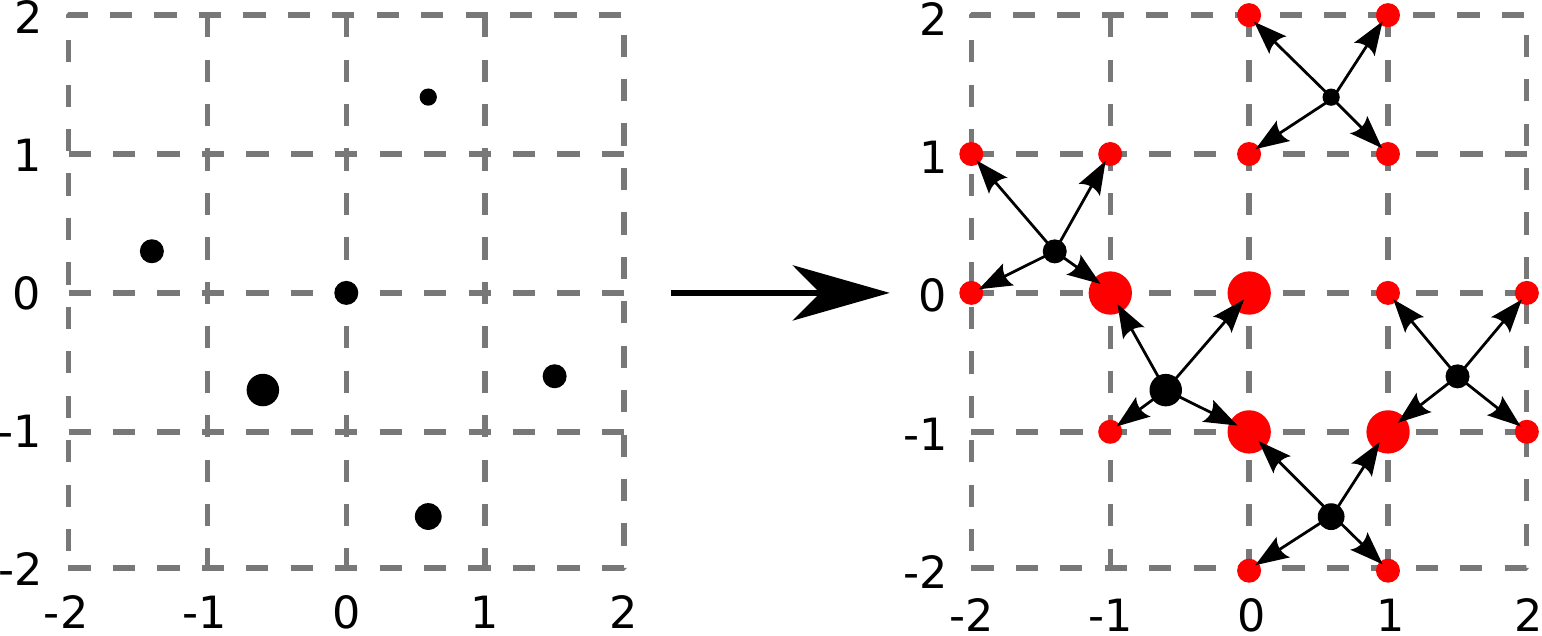}
	\caption{Extrapolation of a filter with multiple synapses. The black dots represent the position of synapses and the red dots represent the extrapolated weights. The large red dots represent an accumulation of multiple weights, and the weights without any dots in the right figure are zero.}
	\label{fig:extrapolation of multiple filters}
\end{figure}

This procedure can be generalized to an ACU with multiple synapses (Fig. \ref{fig:extrapolation of multiple filters}). Each weight of a synapse extrapolated to the nearest points, and the extrapolated weights at the same point are summed up. This leads to the construction of one large convolution weight $\boldsymbol{\overline W}_{\theta_p}$, and the original calculation of an ACU turns out to be the same as that of a conventional convolution using this weight. According to this result, any ACU can be converted to a naive convolution, thus leading to the derivation of Lemma~\ref{thm:conv is super set of acu}.

\smallskip
\begin{lemma}
	\label{thm:conv is super set of acu}
	The set containing all operations represented by ACU is the subset of the set containing all operations represented by convolution, i.e., $\boldsymbol{A} \subset \boldsymbol{C}$
\end{lemma}

\begin{thm}
	\label{thm:conv and acu is same space}
	Based on Lemma~\ref{thm:conv is subset of acu} and Lemma~\ref{thm:conv is super set of acu}, The set containing all operations represented by ACU is equal to the set containing all operations represented by convolution, i.e., $\boldsymbol{C} = \boldsymbol{A}$
\end{thm}

This shows that ACU has the same operational span as a convolution, only but has a different representation. Generally, extrapolated weight is sparse, and sparsity depends on the position parameters and the number of synapses. In a naive convolution, the spatial size of the filter should be large even though the weight is sparse, thus requiring more parameters. However, ACU reduces the number of total parameters by using a position set. Therefore, we can say that ACU is an efficient representation of a sparse weight convolution and can expand the size of the receptive field infinitely without exploding the number of the weight parameters.

%

\subsection{Discussion about ACU}
\label{subsec:Discussion of ACU}

In \cite{jeon2017active}, we showed that changing the naive convolution to an ACU improves network accuracy while retaining the similar number of parameters. In this study, we conducted further experiments on the CIFAR-10 dataset \cite{krizhevsky2009learning} to understand and analyze the effect of ACU. The base network is a 29-layer deep residual network using pre-activation bottleneck blocks~\cite{he2016identity}. This network consists of three-stages with a doubling of width after each state, and its template for residual blocks is $\begin{bmatrix}1\times1, 32\\3\times3, 32\\1\times1, 128\end{bmatrix}$. The networks are trained for 64k iterations, and all results are averages of five runs. Please refer to Appendix~\ref{appendix:simple setup} for details. 
The base network acquired 6.16\% test error with 1.24M parameters (Table~\ref{table:comparison with large conv}).

\subsubsection{Parameter Efficiency}

\begin{table}[!t]
	\renewcommand{\arraystretch}{1.3}
	\caption{Test error (\%) of networks with a large convolution and ACU. By using a relatively smaller number of parameters, ACU achieved a similar result with a large convolution.}
	\label{table:comparison with large conv}
	\centering
	\begin{tabular}{c|c|c}
		\hline
		\textbf{Network}	& \textbf{Test Error(\%)}	& \textbf{Params} \\	\hline \hline
		3$\times$3 Conv	& 6.16 $\pm$ 0.15	& 1.24M\\	\hline
		3$\times$3 ACU	& 5.56 $\pm$ 0.13	& 1.24M\\	\hline 
		7$\times$7 Conv	& 5.49 $\pm$ 0.12	& 3.82M\\	\hline			
	\end{tabular}
\end{table}

To verify the effect of ACU, we changed 3$\times$3 convolutions in residual blocks to an ACU, and the total number of parameter were almost the same. For training the ACU, we used a normalized gradient for position parameters and conducted warming-up of 10k iterations, following our previous work. This simple change can reduce error by 0.6\% (Table~\ref{table:comparison with large conv}).

\begin{figure}
	\centering
	\includegraphics[width=0.4\textwidth]{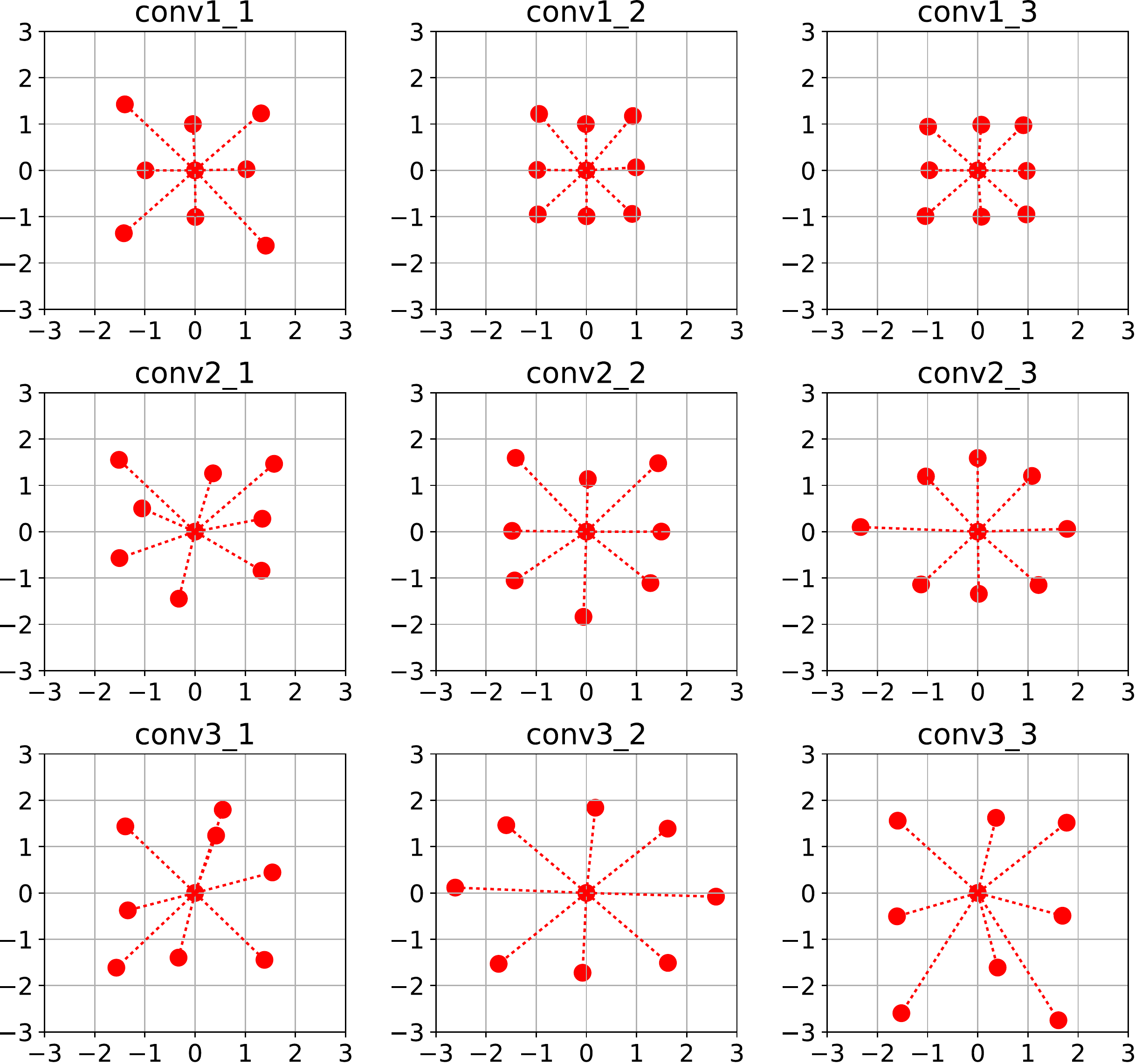}
	\caption{Example of learned positions of ACUs. Higher layers tend to grow its receptive field. Sizes of filters in all ACUs are bounded in the 7$\times$7 area.}
	\label{fig:position_simple_exp}
\end{figure}

Fig~\ref{fig:position_simple_exp} shows the learned positions of ACUs, and the displacement of all synapses are bounded below three. Therefore, as we proved earlier, this network is the subset of the network using 7$\times$7 convolutions. To compare this with the ACU network, we changed 3$\times$3 convolutions to 7$\times$7. Although this network has more than three times the parameters of the ACU network, the performance is similar as in Table~\ref{table:comparison with large conv}. This result shows that the ACU can expand the network capacity with a small number of parameters and optimize the network better than a large convolution. Therefore, we can say that an ACU is an efficient representation of sparse weight convolution.


\subsubsection{Factor of Improvement}
\label{subsubsec:The factor of improvement}
To deeply understand the reason for the improvement brought about by the ACU, we conducted additional experiments, as shown in Table \ref{table:ablation study of ACU}. Fig~\ref{fig:position_simple_exp} shows that high layers tend to enlarge the fields of reception, and this is in agreement with the results in our previous work. Therefore, one might think that the improvement is brought about by the enlargement of the receptive fields in higher layers, and we changed 3$\times$3 convolutions in the last stage to dilated convolutions~\cite{chen2014semantic,YuKoltun2016} with dilation 2 to enlarge its receptive field while retaining the number of parameters (Base-D2). However, this result is even worse than that of the base network, showing that the enlargement of the receptive field, by itself, is not helpful for the network.

If, as claimed earlier, the ACU is able to learn its shape through training, we must determine whether the final shape of the convolution is really the effectively optimized shape for the application. To answer this question, we initialized the shape of ACUs according to the final shape of the convolution of one sample of the ACU network and fixed the shape (ACU-Fixed). Weights were randomly initialized for different runs and trained from their initial state. The result obtained was even further improved from that of the ACU network. This result shows that the shape of the convolution learned by the ACU was optimized for the given application. We believe that this is the main factor of the improvement using ACU. 


\begin{table}[!t]
	\renewcommand{\arraystretch}{1.3}
	\caption{Comparison of test error (\%) with ACU and its variants. Base-D2 applied a convolution with dilation 2 in the last stage. ACU-Fixed was trained using randomly initialized weight with the fixed pretrained shape. These results show that the trained shape using ACUs is effective.}
	\label{table:ablation study of ACU}
	\centering
	\begin{tabular}{c|c|c}
		\hline
		\textbf{Network}	& \textbf{Test Error(\%)}	& \textbf{Improvement(\%)} \\	\hline \hline
		Base & 6.16 $\pm$ 0.15	& -\\	\hline			
		Base-D2	& 6.28 $\pm$ 0.09	& -0.12\\	\hline									
		ACU & 5.56 $\pm$ 0.13	& +0.6\\	\hline
		ACU-Fixed	& 5.44 $\pm$ 0.14	& +0.72\\	\hline			
	\end{tabular}
\end{table}

\section{Grouped Active Convolution}\label{sec:Grouped ACU}

In our previous work~\cite{jeon2017active}, we shared the position set for all output channels in a layer, and each layer has only one shape. To view multiple receptive fields for ACU, we might apply different shapes to some output channels. However, in this approach, every input channel should be interpolated for all other shapes, whereas the interpolation for the input is required only once in the original ACU. This leads to numerous calculations, thus slowing down the process. Moreover, no significant improvement could be found to offset this overhead. Therefore, the practical application of multiple shapes for an ACU has not been straightforward.

ResNeXt~\cite{xie2017aggregated} showed that convolution with grouping is more efficient than a naive convolution. A grouped convolution divides input channels according to groups and calculates the corresponding output by using only the input channels within the same group. This saves lots of weights for a convolution. Recently, many studies~\cite{zhang2017shufflenet, chollet2016xception, howard2017mobilenets, sandler2018inverted} used grouped convolution for improving the accuracy and reducing the number of parameters. Xie et al.~\cite{xie2017aggregated} denoted the number of groups as a cardinality; we used the same terminology in this study. 

Motivated by the grouped convolution, we propose the grouped ACU, which enables to use multiple position parameter sets without increasing computation complexity. The grouped ACU shares position parameters within a group and uses different parameter sets for different groups; therefore, the number of position sets is the same as the cardinality. Eq.~\eqref{eq:position param} can be extended to Eq.~\eqref{eq:multiple position params} in a grouped ACU.

\begin{equation}
\begin{aligned}
\label{eq:multiple position params}
&\theta_P = \bigcup_{1\leq g \leq G}\theta_p^g ,\\
&\theta^g_p = \{p^g_k|0\leq k < K\},\\
&p^g_k = (\alpha^g_k, \beta^g_k) \in \mathbb{R}^2
\end{aligned}
\end{equation}
where $G$ is the cardinality and $K$ is the number of synapses, respectively. In addition, $\theta^{g}_{p}$ is the set of position parameters for each group $g$.

\begin{figure}
	\centering	
	\begin{tabular}{p{0.45\columnwidth}p{0.45\columnwidth}}
		\multicolumn{2}{c}{\includegraphics[width=0.9\columnwidth]{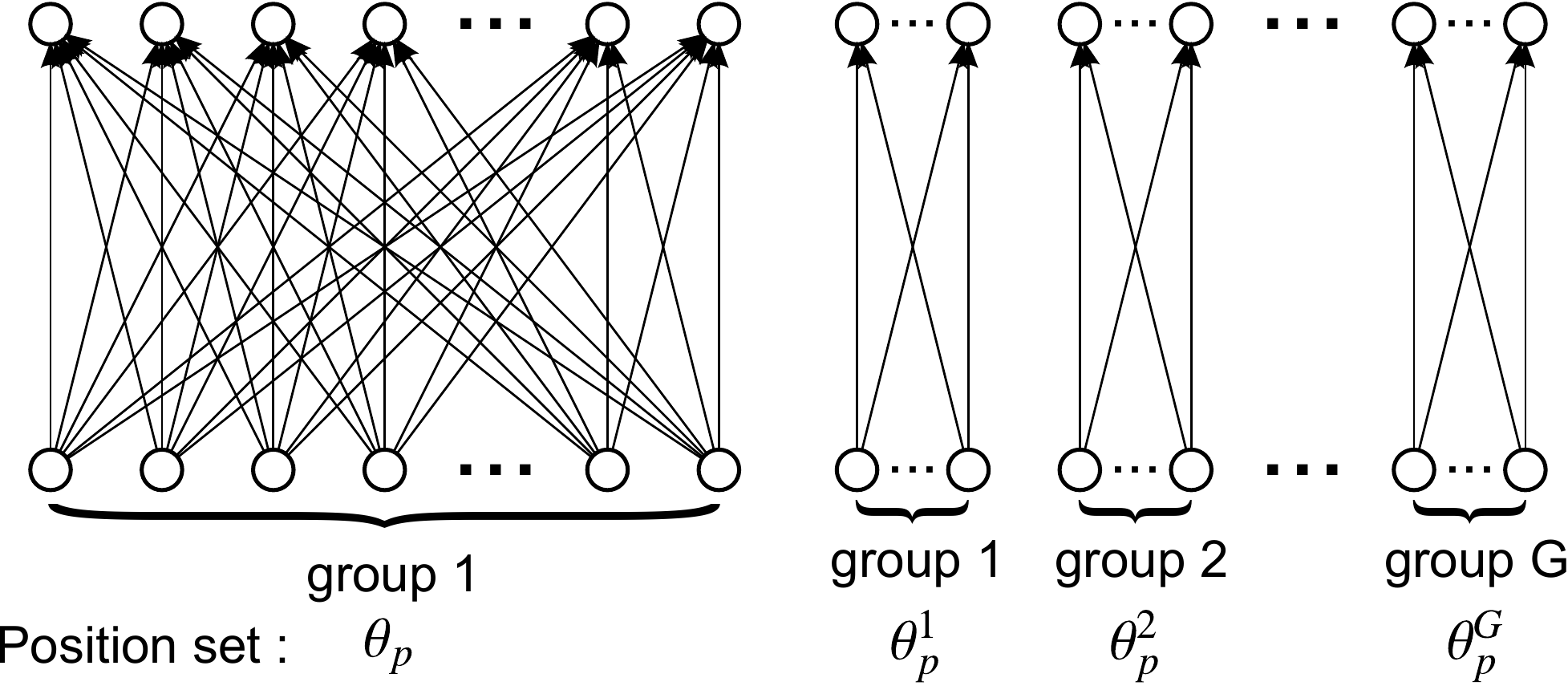}}\smallskip\\
		\centering (a)ACU & \centering (b)Grouped ACU		
	\end{tabular}	
	\caption{(a) ACU connects all input and output channels and uses only one position set, which is shared among all output channels. (b) Grouped ACU splits the input and output channels according to the groups, and links channels only within the same groups. Each group has its own position set, which is optimized only for that group. The upper and lower circles represent channels.}
	\label{fig:grouped ACU}
\end{figure}

Figure~\ref{fig:grouped ACU}(b) shows the concept of the grouped ACU. Compared to the original ACU, different position sets are applied to different groups in the grouped ACU. As this grouped ACU divides input channels according to groups, each input is interpolated only once regardless of the cardinality; therefore, the computational complexity for interpolation is the same as that of the original ACU. The total computational complexity is decreased owing to the reduction in the linkage from input to output channels. With the grouped ACU, convolutions in each group possess their individual shapes, and multiple receptive fields can be observed in a layer. The ACU we proposed in our earlier work can be regarded as a grouped ACU with only one group.

The number of the weight parameters for the grouped ACU is the same as that for grouped convolution. The number of the additional parameters for a grouped ACU is $2\times(K-1)\times G$; each synapse has two parameters for defining positions, and position parameters are not required at the origin because its position is fixed. As a result, the total number of parameters for the grouped ACU, except bias, includes

\begin{equation}
\label{eq:num of params}
\{\frac{C_I}{G} \times \frac{C_O}{G} \times K \times  G\} + \{2\times(K-1)\times G\}
\end{equation}

where $C_I$ and $C_O$ are the numbers of input and output channels, respectively. Generally, as the widths of channels $C_I$ and $C_O$ are greater than the number of synapses $K$, if cardinality $G$ is small, the first term is considerably larger than the second term. With the increase in cardinality, the second term also increases but the total number of parameters decreases as the first term is dominant.

\begin{figure*}
	\centering
	\begin{tabular}{p{0.24\textwidth}p{0.24\textwidth}p{0.22\textwidth}p{0.18\textwidth}}
		\multicolumn{4}{l}{\includegraphics[width=0.92\textwidth]{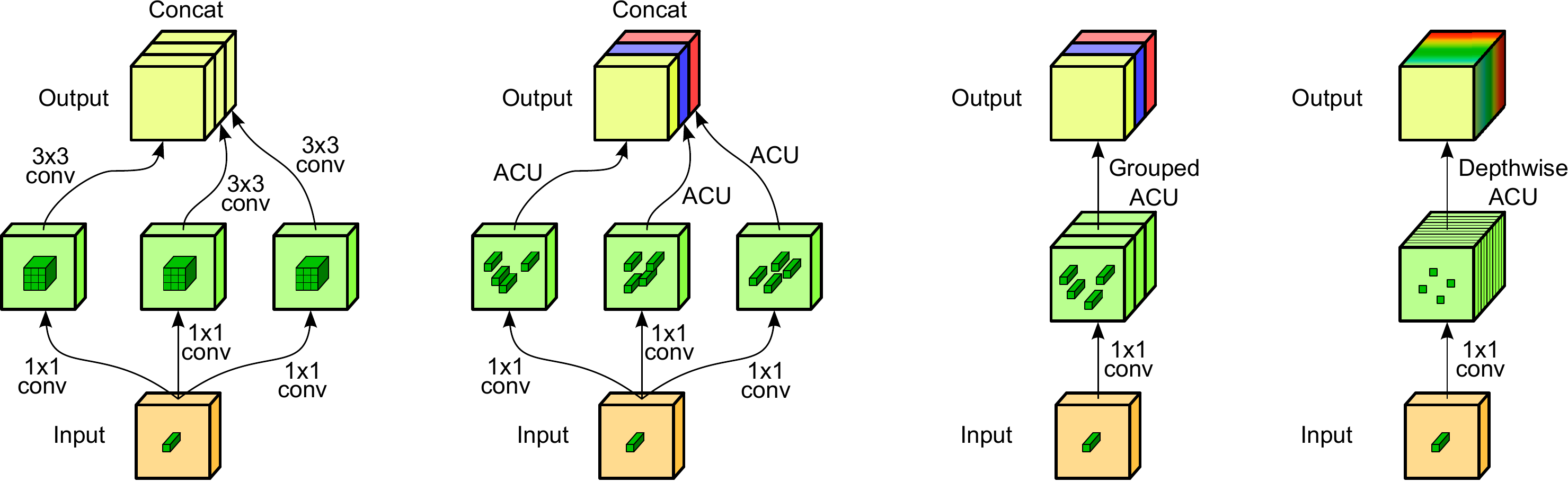}} \smallskip\\		
		\centering\shortstack{(a) Simplified Inception\\ \phantom{}} & 
		\centering\shortstack{(b) Generalized Inception\\based on ACUs} & 
		\centering\shortstack{(c) Generalized Inception\\based on grouped ACU } &
		\shortstack{(d) Extream Inception\\ with depthwise ACU }
	\end{tabular}
	\caption{Generalization of Inception structures: (a) Simplified Inception with 3$\times$3 convolutions proposed in \cite{chollet2016xception}. This simplification only combines single-scale features. (b) In the generalization of Inception through ACUs, each ACU views different receptive fields. (c) Equivalent conversion of (b) through grouped ACU. (d) Extension of Inception to the extreme case through depthwise ACUs, which observes different receptive field for every input channel. Tensors of different colors represent output channels with the application of different receptive fields. (Best viewed in color)}
	\label{fig:inception struct}
\end{figure*}

\subsection{Generalization of Inception}
\label{sec:GACU is generalized Inception}
The Inception module~\cite{szegedy2015going} is composed of multiple types of convolutions and poolings, and achieves good accuracy while
maintaining a similar computational budget. The key feature of this success is based on the composition of multiple receptive-field convolutions. However, it is difficult to design such an Inception structure because its components and parameters are carefully and manually decided by considering questions such as how many different types of convolutions should be used, what type of convolution should be used, and how many channels must be applied for convolutions.

Chollet~\cite{chollet2016xception} simplified Inception by using 3$\times$3 convolutions (Fig.~\ref{fig:inception struct}(a)). However, this simplification is not general enough to reflect the key aspect of an Inception module because it has only one receptive field. The important characteristic of Inception is the use of multiple receptive fields. When we change 3$\times$3 convolutions in Fig.~\ref{fig:inception struct}(a) to an ACU (Fig.~\ref{fig:inception struct}(b)), this structure can receive multiple receptive fields as each ACU is able to learn its own shape through training. Therefore, we claim that this structure can be a more general form of the Inception module. Fig. \ref{fig:inception struct}(b) is equivalently represented by a grouped ACU following a 1 $\times$ 1 convolution (Fig.~\ref{fig:inception struct}(c)). This generalization based on the grouped ACU significantly simplifies the design of an Inception module while still taking advantage of the key characteristic of the Inception, namely having multiple receptive fields; only the width and cardinality of a grouped ACU should be selected, and an ACU will automatically select multiple receptive fields through training.

By further increasing the cardinality, we reached an extreme case, in which only one input channel is used for producing output. This kind of convolution is a special case of the grouped convolution and is called depthwise convolution. Fig.~\ref{fig:inception struct}(d) shows the depthwise ACU, which applies the ACU to the depthwise convolution. In this unit, each output channel has its own shape, and it accepts as many receptive fields as the number of channels available.

\subsection{Effectiveness of Grouped ACU}
\label{subsec:cardinality vs performance}

\begin{table*}[!t]
	\renewcommand{\arraystretch}{1.3}
	\caption{Comparison of test error (\%) with the increase in cardinality. The results are plotted in Fig. \ref{fig:error vs group} }
	\label{table:ex grouped ACU}
	\centering
	\begin{tabular}{c|cc|cc|cc}
		\hline
		& \multicolumn{2}{c|}{\textbf{Conv3$\times$3}} & \multicolumn{2}{c|}{\textbf{Conv5$\times$5}} & \multicolumn{2}{c}{\textbf{ACU}}\\ 					
		\textbf{Network}	& \textbf{Test Error(\%)}	& \textbf{Params(M)}
		& \textbf{Test Error(\%)}	& \textbf{Params(M)}
		& \textbf{Test Error(\%)}	& \textbf{Params(M)}\\	\hline \hline
		16$\times$4d & 5.38 $\pm$ 0.06	& 1.28
		& 4.78 $\pm$ 0.1	& 1.54
		& 4.93 $\pm$ 0.14	& 1.28\\	\hline	
		
		$\times$4d & 5.43 $\pm$ 0.13	& 1.18		
		& 4.96 $\pm$ 0.14	& 1.27		
		& 4.87 $\pm$ 0.19	& 1.19\\	\hline	
		
		$\times$2d & 5.59 $\pm$ 0.06	& 1.16
		& 5.06 $\pm$ 0.13	& 1.2
		& 4.83 $\pm$ 0.16	& 1.17\\	\hline	
		
		$\times$1d & 5.75 $\pm$ 0.17	& 1.15
		& 5.18 $\pm$ 0.19	& 1.17
		& 4.9 $\pm$ 0.08	& 1.17\\	\hline	
	\end{tabular}
\end{table*}

To prove the effectiveness of the grouped ACU, we conducted experiments (Table \ref{table:ex grouped ACU}). The base network is similar to the 29-layer deep residual network defined in Section~\ref{subsec:Discussion of ACU}. For utilizing benefits of using grouped convolutions, we converted the residual blocks of the base network to $\begin{bmatrix}1 \times 1, 64 & \\3 \times 3, 64, & 
G=16\\1 \times 1, 128& \end{bmatrix}$ and denote this network as 16$\times$4d following ResNeXt~\cite{xie2017aggregated}, which means the cardinality is 16 and there are four input channels per group in the first stage; this model keeps cardinality constant for all stage and the number of input channels per group is doubled after each stage. This modification reduces test error compared to the network using naive convolutions in Section~\ref{subsec:Discussion of ACU} with a similar number of parameters. 

Firstly, we simply changed the grouped convolutions to grouped ACUs on 16$\times$4d network. As the cardinality is 16, ACUs have 16 different position sets. This network achieved 4.93\% error rate, which shows a 0.45\% improvement compared to the network with naive grouped convolutions. After that, we increased cardinality to examine the effect of using more multiple receptive fields. ResNeXt maintains the cardinality constant regardless of a stage, and thus the number of input channels in a group is doubled after a stage. We changed this model slightly such that the new model retains the number of input channels per group throughout all the stages instead of retaining cardinality. The purpose of this modification is to include an interesting case where all ACUs in the three stages are depthwise ACUs. In this modification, the cardinality is doubled after each stage. As we keep the number of input channels per group constant, we denote this network as $\times$4d (Table \ref{table:ex grouped ACU}), which retains four input channels per group for all stages. The cardinality of the $\times$4d network is the same as that of 16$\times$4d network for the first stage. However, at the second stage, the cardinality reaches 32. This network has the smaller number of parameters compared with 16$\times$4d because the linkage from input to output channels becomes sparser at the second and third stages.

\begin{figure}
	\centering
	\includegraphics[width=0.45\textwidth]{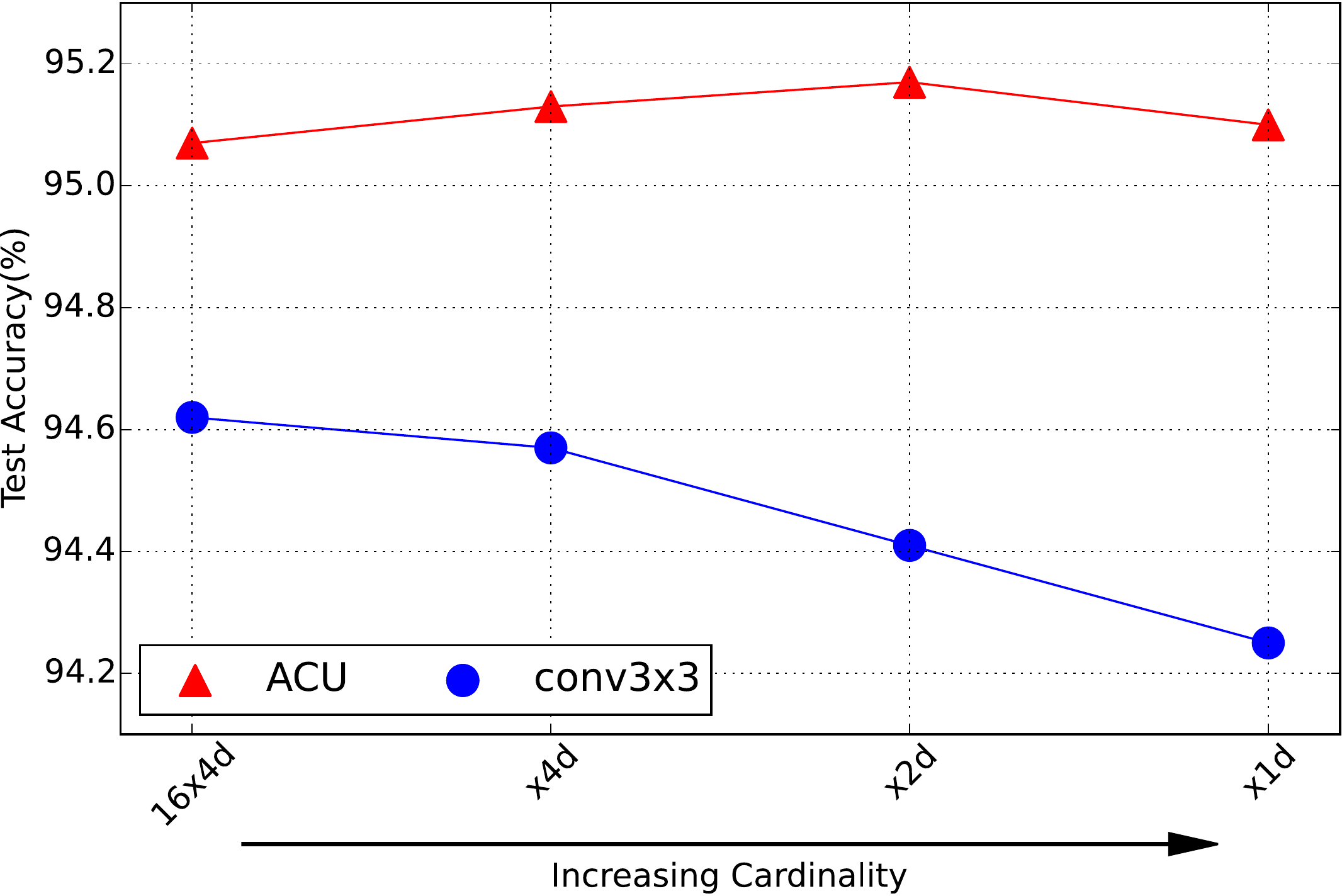}
	\caption{Test accuracy of networks in Table \ref{table:ex grouped ACU}. By increasing cardinality, the accuracy of the network using naive convolution is degraded. However, the network using an ACU retains the accuracy even though the number of parameters decreases.} 
	\label{fig:error vs group}
\end{figure}

We tested with the number of input channels per group being 4, 2, and 1 (Table \ref{table:ex grouped ACU}); in other words, we increased the cardinality. As the cardinality increases, the number of parameters decreases. Fig. \ref{fig:error vs group} summarizes the experimental results, showing an interesting phenomenon. That is, if we use naive convolution, the accuracy is decreased with the decrease in the number of channels per group. This shows that an increase in the cardinality is not always helpful, and a grouped convolution possesses an optimal cardinality. The reduction in the number of channels per group implies that the available feature maps for convolution in each group are reduced, and naive convolution thus forms more limited combinations with a small number of channels in each group. This is why increasing cardinality decreases accuracy.




\begin{figure}
	\centering
	\includegraphics[width=0.9\linewidth]{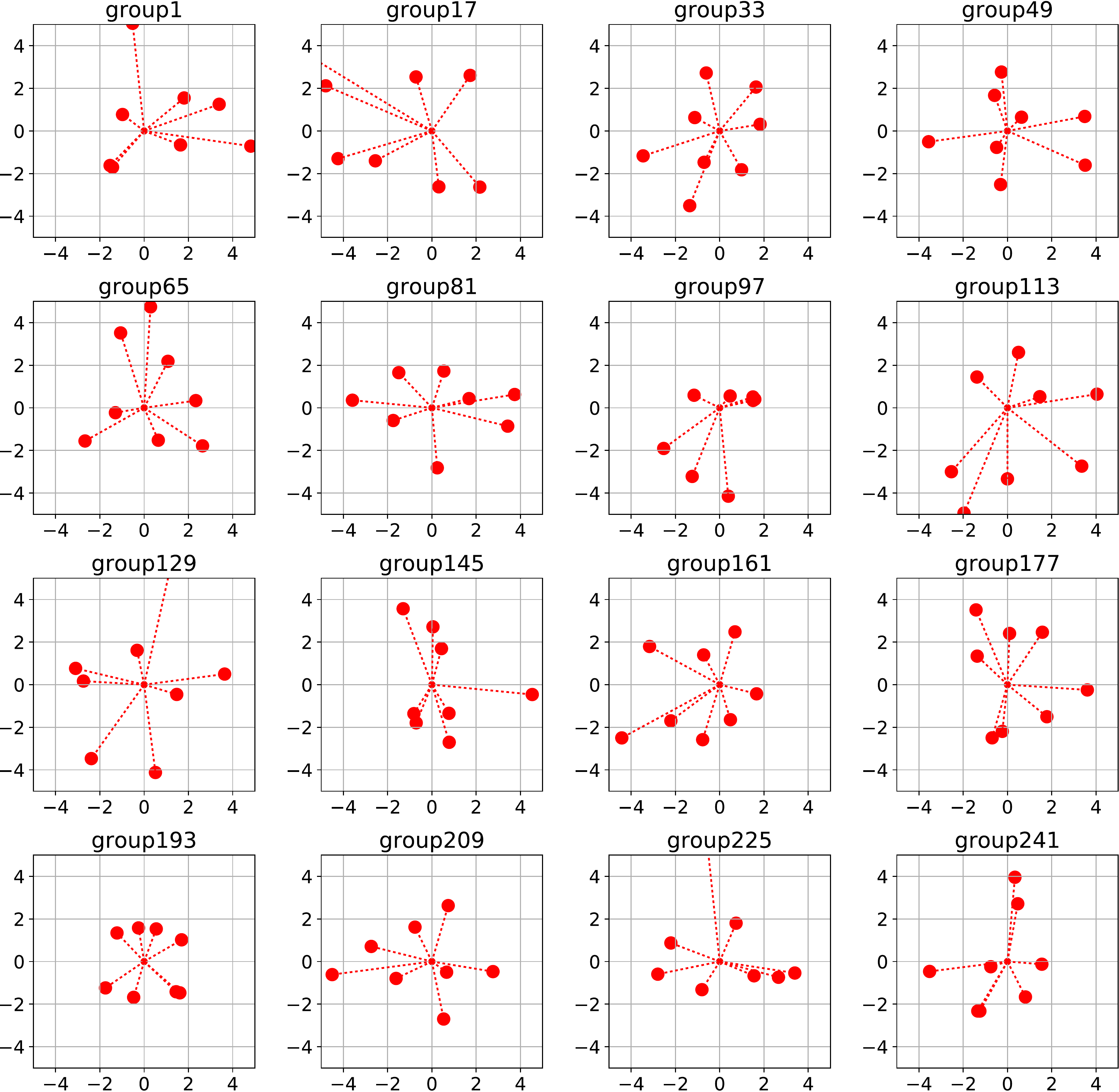}
	\caption{Example of learned position in one layer (ACU in the last residual block of $\times$1d network in Table \ref{table:ex grouped ACU}). This shows diversity of learned positions.} 
	\label{fig:sample of learned position of grouped ACU}
\end{figure}

\begin{figure*}
	\centering	
	\begin{tabular}{p{0.4\textwidth}p{0.4\textwidth}}
		\multicolumn{2}{c}{\includegraphics[width=0.8\textwidth]{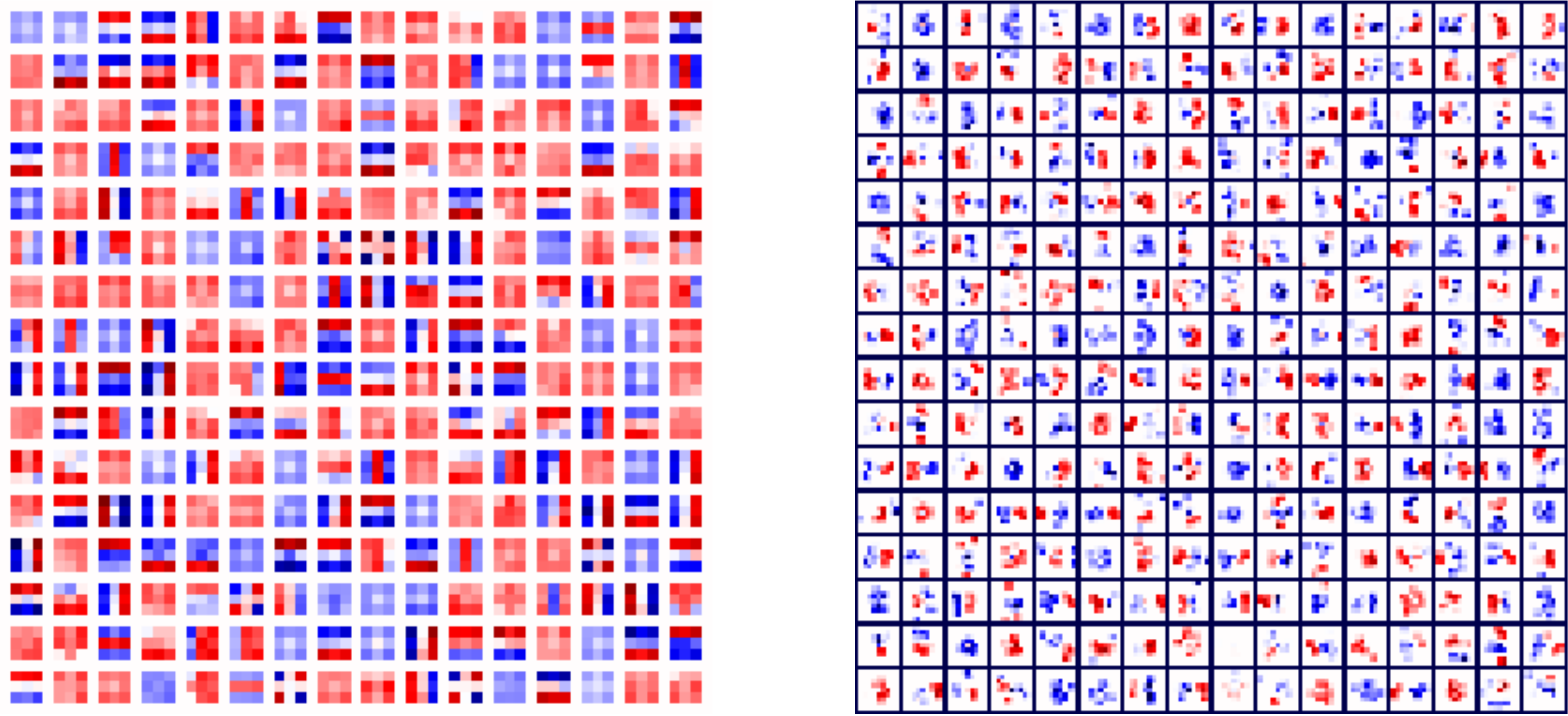}}\smallskip\\
		\centering (a) Filters of 3$\times$3 convolution& \centering ~~(b) Extrapolated filters of ACU
	\end{tabular}	
	\caption{Visualization of the filter of the 3$\times$3 convolution in the last residual block of $\times$1d network in Table \ref{table:ex grouped ACU} (a) Owing to the restriction of its shape, many duplications of filters are shown in naive convolution. (b) The filters are shown as diverse shapes and display many white areas indicating sparsity. To visualize the filter of ACU, we extrapolated filters and cropped a 9$\times$9 area. Note that blue and red represent negative and positive values, respectively.}
	\label{fig:Filter_visualization}
\end{figure*}

In contrast, in a grouped ACU, an increasing cardinality does not degrade the accuracy. In Fig. \ref{fig:error vs group}, the error rate remains almost the same while the number of channels per group decreases. Naive convolution has a fixed shape, because of which the way input feature maps are combined spatially and across channels is quite limited. However, the corresponding ACU could expand its receptive field and the combination is not so restricted even though the number of channels per group is decreased due to the increased freedom in the shape of the receptive field. Fig. \ref{fig:sample of learned position of grouped ACU} shows an example of the ACU-learned shapes of convolutions in a layer. The figure also shows a diversity of receptive fields. In addition, Fig.~\ref{fig:Filter_visualization} shows the visualization of filters (the filter of 3$\times$3 convolution in the last residual block of $\times$1d network in Table \ref{table:ex grouped ACU}). While we can see many duplications in naive convolution, the filter of ACU is more diverse and widely spread.

Actually, the degree of freedom for one filter in a grouped ACU is larger than that in a naive convolution owing to the additional position parameters. Therefore, we can suspect that the characteristics, which maintain performance with a smaller number of input channels per group, originate from a larger number of free parameters. To complete our analysis, we conducted further experiments by changing 3$\times$3 convolutions to 5$\times$5 convolutions, which results in more or at least the same number of parameters than the 3$\times$3 ACU (with $K=9$) network as shown in Table \ref{table:ex grouped ACU}. We also observed the same tendency with 5$\times$5 convolutions, i.e., with the decreasing number of input channels in a group, the accuracy is also decreased. A depthwise 5$\times$5 convolution, which has $5\times5=25$ parameters per group, has exactly the same number of parameters as that in a 3$\times$3 depthwise ACU, which has $3\times3+2\times(9-1)=25$ parameters; however, the accuracy of this network is worse than that of ACU.

Therefore, the benefit of using an ACU to develop useful features by using only a small number of input channels does not originate from the number of parameters but from its ability to change its shape. Considering that $\times$1d has a smaller number of parameters than $\times$4d, $\times$1d is a good choice for the grouped ACU; thus, we used the depthwise ACU for the remainder of this study.

\subsection{Discussion}

\subsubsection{Effect of Warming up}
In \cite{jeon2017active}, we suggested the strategy of ''warming-up," which freezes the position parameters for some iterations to stabilize the initial weight; this showed an additional performance boost. However, in the depthwise ACU, we found out that warming-up is not needed and is rather harmful. Table \ref{table:effect of warming up} shows the results of test accuracy with and without warming-up. When we use one group of ACU, the stability of its shape is important because the position parameter is shared among all channels. Therefore, the early movement of position with unstable weight is not good for the training, and warming-up of the weight is a beneficial option to reduce this effect.

However, in a depthwise ACU, each input--output channel pair does not share position parameters but uses its own set, and the position is optimized according to the initial weight. Thus, the early movement of position is not an unstable behavior. In contrast, warming-up prevents the movement of the position and restricts diversity in shapes. In this analysis, we did not use warming-up for depthwise ACUs.

\begin{table}
	\renewcommand{\arraystretch}{1.3}
	\caption{Test error (\%) with and without warming-up. If ACU shares positions for all outputs, warming-up is helpful. However, in a depthwise ACU, the effect of warming-up is conflicting.}
	\label{table:effect of warming up}
	\centering
	\begin{tabular}{c|c|c|c}
		\hline
		\textbf{Network}	& \textbf{Baseline}	& \textbf{With warming-up}  & \textbf{Improvement}  \\	\hline \hline
		ResNet/ACU & 5.64 & 5.56 & +0.08\\	\hline
		$\times$1d/ACU & 4.9 & 5.04 & -0.14\\	\hline
	\end{tabular}
\end{table}

\subsubsection{Effectiveness of Using Multiple Positions}
A depthwise ACU uses the same groupings for convolution and position parameters. However, conceptually, we can use two different groupings for computing convolutions and sharing positions. For example, it is possible to use four groups for convolutions but two groups for positions: groups 1 and 2 for convolution share the same position parameters, while groups 3 and 4 share the other set of position parameters.

If a grouped ACU is applied, the improvement can come from the ACU itself and/or by using multiple positions of the ACU. To clarify the factor for the improvement, we examined another type of ACU, which operates similar to the depthwise ACU but uses only one set of position parameters and shares them for all channels. This implies that a layer has one receptive field regardless of the cardinality.

Even with the shared receptive field ($\times$1d/ACU/Share in Table~\ref{table:effect of multiple positions}), a better result was achieved than that for the base network. This shows that the ACU is effective even though using only one set of position parameters. If we use multiple positions ($\times$1d/ACU/Multi), we can obtain additional improvement; this supports the effectiveness of using multiple position parameters.

\begin{table}[!t]
	\renewcommand{\arraystretch}{1.3}
	\caption{Effect of using multiple positions for depthwise ACUs. By using multiple positions ($\times$1d/ACU/Multi), we can achieve additional improvement compared to that obtained using only one position parameter set ($\times$1d/ACU/Share)}
	\label{table:effect of multiple positions}
	\centering
	\begin{tabular}{c|c|c}
		\hline
		\textbf{Network}	& \textbf{Test Error (\%)}	& \textbf{Params} \\	\hline \hline
		$\times$1d 			& 5.75 $\pm$ 0.17	& 1.15M\\	\hline	
		$\times$1d/ACU/Share	& 5.2 $\pm$ 0.04	& 1.15M\\	\hline	
		$\times$1d/ACU/Multi 		& 4.9 $\pm$ 0.08	& 1.17M\\	\hline	
	\end{tabular}
\end{table}

\subsubsection{Retraining with Trained Shapes}
In Section~\ref{subsubsec:The factor of improvement}, we achieved better results by retraining the network using trained position parameters on naive ACU than training it from scratch. We can consider applying the same strategy to the case of depthwise ACU network. However, it turned out that this strategy does not work well with depthwise ACU. An ACU with one group shares position parameters for all output channels; thus, the optimized shape generally works well for that layer. Therefore, the trained shape can also be effective for reinitialized weight. However, in a depthwise ACU, each position set is specialized only for a particular weight. If we reinitialize the weights with an optimized shape, the weight would not match with the trained shape. As a result, obtaining a better accuracy than that of the original network is difficult. Nevertheless, we conducted experiments, and as expected, obtained worse results; the test accuracy is decreased by 0.38\%.

%


\section{Experiment}\label{sec:experiment}

\subsection{Experiment on ImageNet}
We conducted experiments on the ImageNet classification task\cite{ILSVRC15} on several networks to show the effectiveness of the proposed unit. To observe mainly the effect of ACU itself, we used a single set of hyperparameters and did not apply intensive data augmentation. We randomly cropped 224$\times$224 from 256$\times$256 images based on the method in \cite{Krizhevsky2012} and flipped the images horizontally. The inputs were normalized, and no more augmentations were applied. To reduce the training time, we used a linear decay learning-rate schedule with 60 epochs. The initial learning rate was 0.1. An extensive search was not conducted for finding the optimal hyperparameters for the given networks. All the results were derived using a single network and performing single crop testing on a validation set, with an average of three runs.

\subsubsection{Replacing Inception Modules}
In Section~\ref{sec:GACU is generalized Inception}, we claim that a grouped ACU can generalize Inception modules, and we conducted the experiment to support this idea. We started with an Inception network~\cite{szegedy2015going} and applied BN~\cite{ioffe2015batch} before all ReLUs to speed up the training. This is the base network and we achieved 72.0\% Top-1 accuracy (Table \ref{table:result of Inceptions}). Then, from the base network, we simply changed all of the Inception modules to a (1$\times$1 conv)-(3$\times$3 depthwise ACU) block like in Fig.~\ref{fig:inception struct}(d). When we retain the width of the Inception, this conversion reduces the number of parameters. To compare the performance by using a similar parameter size, we expanded the width of our blocks by a factor of 1.4. 

\begin{figure}
	\centering
	\includegraphics[width=0.9\linewidth]{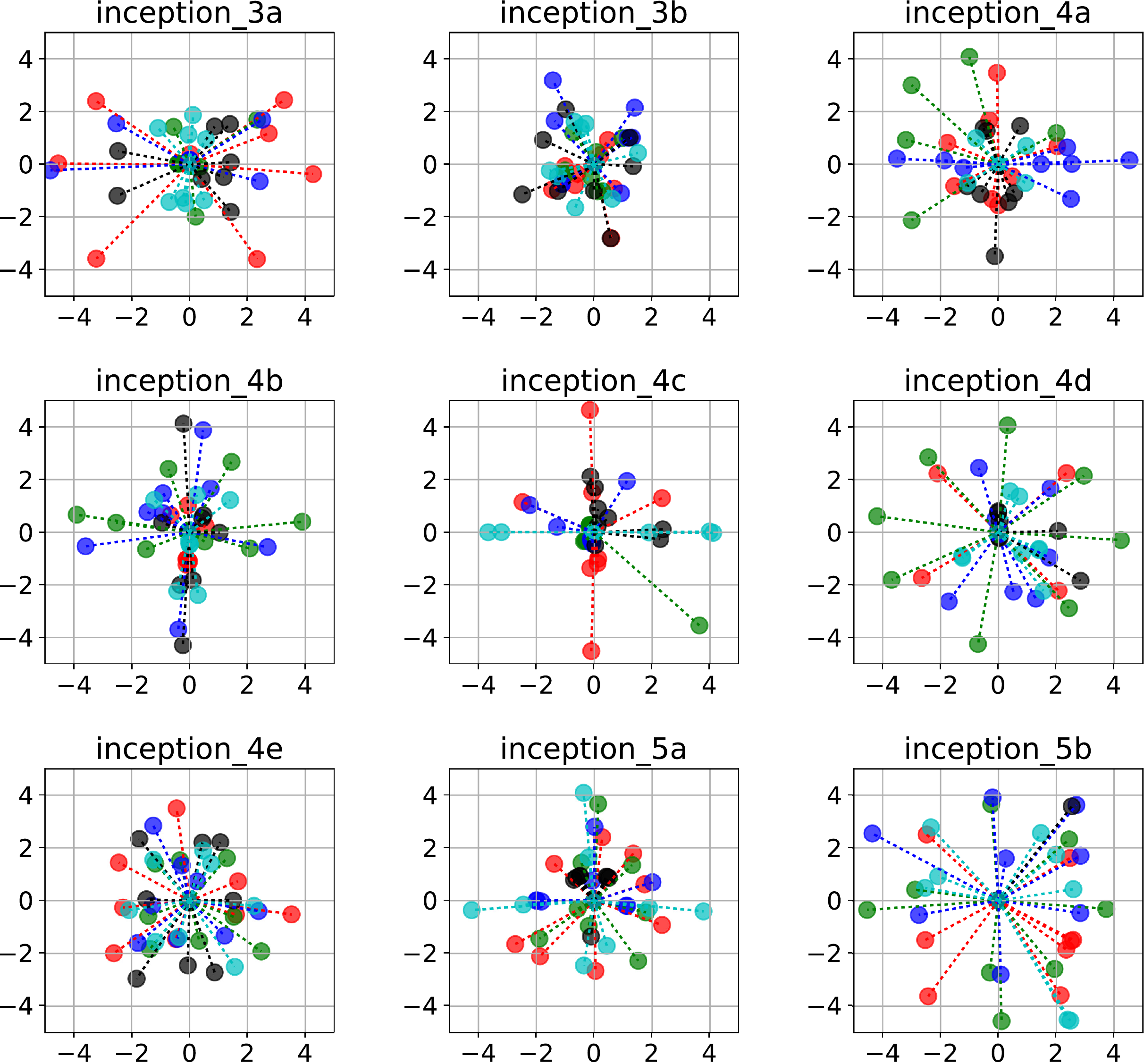}
	\caption{Learned positions of depthwise ACUs corresponding Inception blocks. Five randomly sampled position sets of each ACU layer are drawn. Multiple receptive fields are applied to each channel. Each color represents each position set. (Best viewed in color).}
	\label{fig:inception_acu_position}
\end{figure}

This change reduces the computational complexity in terms of the total number of Multiply--Adds (MAdds), due to the simplification of Inception modules. Further, this network achieved a slightly better result compared to that using the original method (Table \ref{table:result of Inceptions}). Fig. \ref{fig:inception_acu_position} shows the samples of the learned positions of each ACU corresponding Inception block. Five channels were selected randomly and their position sets are represented by different colors. Diverse shapes of convolutions can be observed; therefore, multiple receptive fields can be applied in a layer like Inception module. This new block is simple and is not needed to decide which type of convolution or pooling should be used or how many channels must be assigned for each operation. This simplification can help develop effective architectures by reducing the complexity of design choices.

\begin{table}[t!]
	\renewcommand{\arraystretch}{1.3}
	\caption{The effectiveness of replacing Inception blocks. This is the Top-1 accuracy (\%) on ImageNet dataset. If we change Inception blocks to our blocks proposed in Fig.~\ref{fig:inception struct}(d), a better result with less computation complexity can be achieved, and our block is much simpler than original Inception block.}
	\label{table:result of Inceptions}
	\centering
	\begin{tabular}{c|c|c|c}
		\hline
		\textbf{Network}	& \textbf{Accuracy}	& \textbf{Params} & \textbf{MAdds}
		\\	\hline \hline
		Inception\cite{szegedy2015going}+BN
		& 72.0	&  7.0M & 1.58G\\	\hline
		Replace Inception blocks to ours
		& 72.1	&  7.1M & 1.47G\\	\hline		
	\end{tabular}
\end{table}

\subsubsection{Xception with Depthwise ACUs}
Xception~\cite{chollet2016xception} is considered as an extreme version of the Inception module using depthwise convolutions. However, as stated previously, Xception has replaced Inception block with only 3$\times$3 convolutions and this is not enough to generalize Inception because such a replacement would not be able to show multiple receptive fields, which is an important characteristic of an Inception block. 

\begin{figure}
	\centering
	\includegraphics[width=0.9\linewidth]{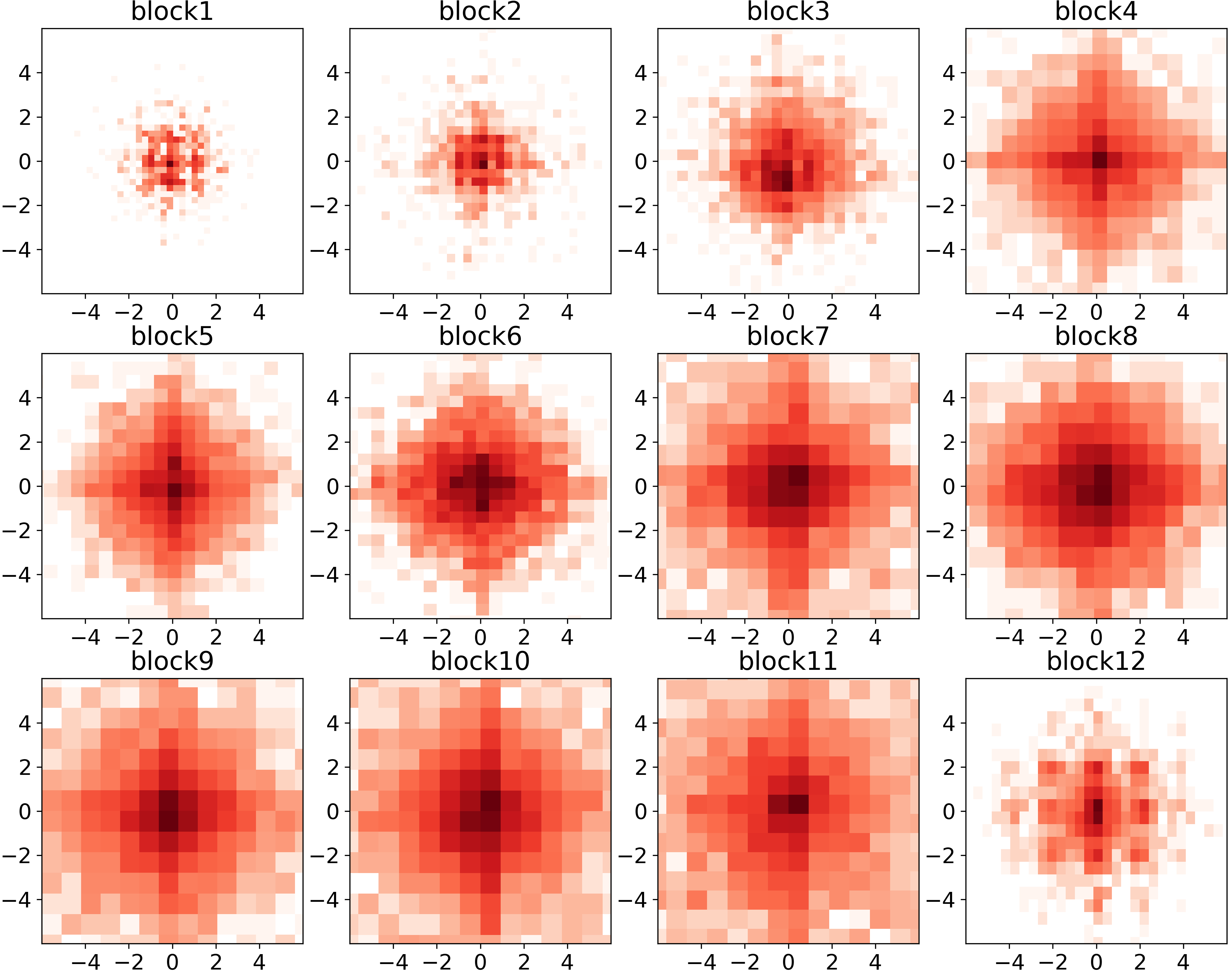}
	\caption{2D Histogram of learned shapes of an depthwise ACUs in Xception network. The darker colors indicate higher frequency (in log scale). The diversity of synapse positions is observed.}
	\label{fig:xception_acu_position_histogram}
\end{figure}

For verifying that the use of a depthwise ACU is more effective than the use of a naive depthwise convolution, we changed the depthwise convolutions to depthwise ACUs. Because we regarded a residual block as one Inception module, we only changed the last depthwise convolution to a depthwise ACU in each residual block. The last depthwise ACU summarizes features in a block by observing multiple receptive fields. This modification resulted in the addition of small amounts of parameters, and we achieved a better result than the base network (Table \ref{table:result of Xception}). Fig.~\ref{fig:xception_acu_position_histogram} illustrates the 2D histogram of synapse positions of ACU in each residual block. The distribution of positions varies according to the depth of a layer. Compared to naive depthwise convolutions, which can view only a fixed area, our unit can view more diverse receptive fields. 

\begin{table}[t!]
	\renewcommand{\arraystretch}{1.3}
	\caption{Comparison of depthwise convolution and depthwise ACU on Xception\cite{chollet2016xception} network. This is the Top-1 accuracy (\%) on ImageNet dataset. As the depthwise ACU can see multiple receptive fields, we achieved a better result on Xception than using naive depthwise convolutions.}
	\label{table:result of Xception}
	\centering
	\begin{tabular}{c|c|c|c}
		\hline
		\textbf{Network}	& \textbf{Accuracy} & \textbf{Params} & \textbf{MAdds}
		\\	\hline \hline
		Xception with depthwise conv
		& 75.6	& 22.9M & 4.6G\\	\hline
		Xception with depthwise ACU
		& 75.9	& 23.0M & 4.7G\\	\hline
	\end{tabular}
\end{table}

\subsubsection{Depthwise ACU for Mobile applications}
Depthwise convolution reduces not only the number of parameters but also the computational complexity. Accordingly, attempts have been made to apply it to mobile applications. MobileNet~\cite{howard2017mobilenets, sandler2018inverted} is an up-to-date network which runs fast in mobile devices maintaining high accuracy. This network uses depthwise convolutions to achieve good accuracy with feasible running time. 

Although the calculation of depthwise ACU is slightly complicated than that of a naive convolution because of the presence of interpolations, the calculation of a depthwise operation utilizes a small portion of the total network computation; the fully connected 1$\times$1 convolution utilizes considerable amounts of computations over the entire network. Therefore, we can think about applying depthwise ACU to mobile applications and examined this potential through experiments. 

\begin{figure}
	\centering
	\includegraphics[width=0.9\linewidth]{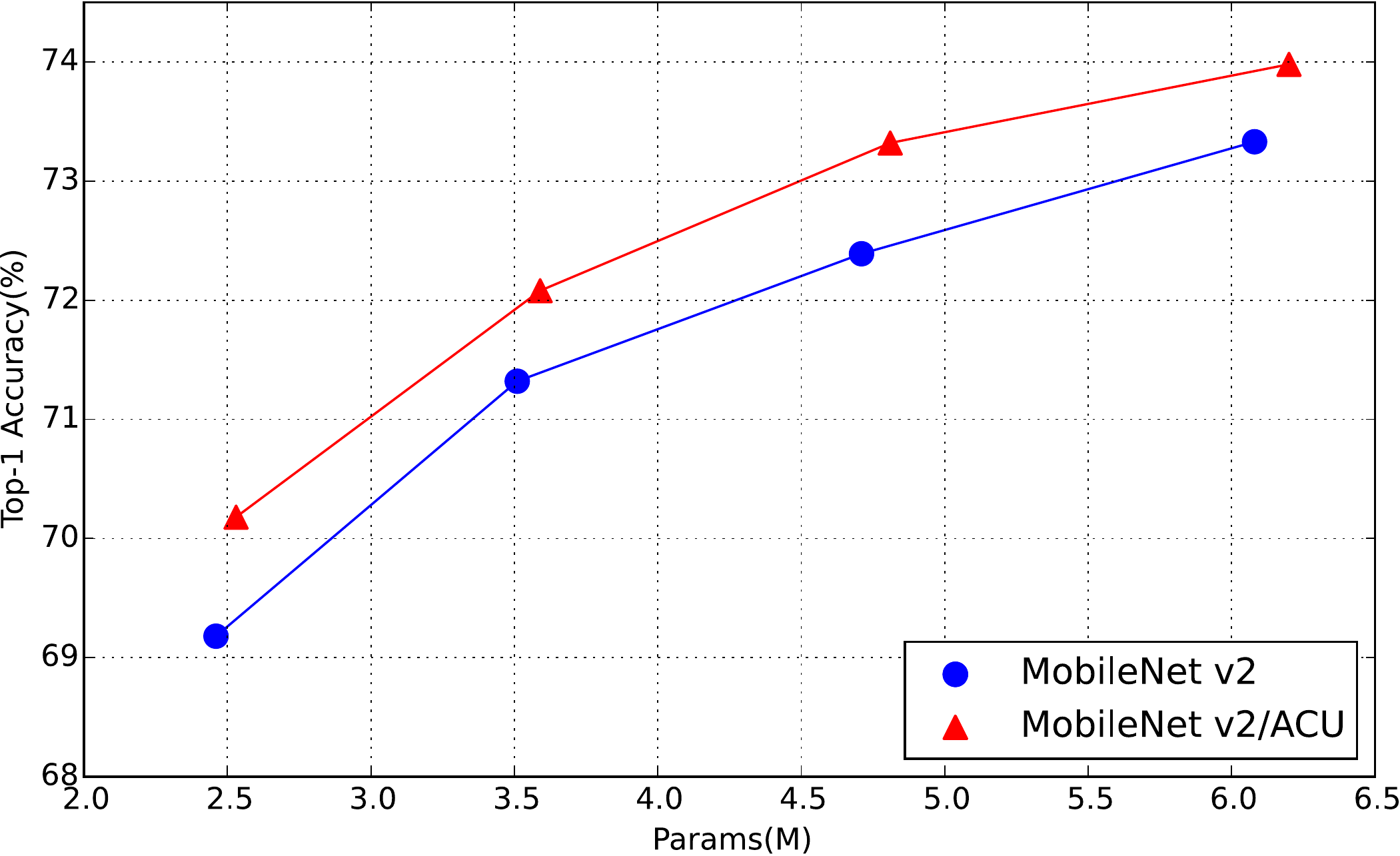}
	\caption{Top-1 accuracy vs. the number of parameters in MobileNet v2~\cite{sandler2018inverted}. With the same number of parameters, the performance of a network using depthwise ACU is consistently better than that of the base network.}
	\label{fig:mobilenet_param_vs_accuracy}
\end{figure}

The base network we used in this study is the MobileNet v2~\cite{sandler2018inverted}, which consists of repetitions of residual blocks with depthwise convolution. We changed all depthwise convolutions in the residual blocks, and Table \ref{table:result of MobileNet} shows the experimental result. We performed tests by varying the width multiplier from 0.8 to 1.4; this controls the whole width of the network. Fig.~\ref{fig:mobilenet_param_vs_accuracy} summarizes the results and shows that the network with ACU consistently achieved better accuracy with the same number of parameters. 

To compare the performance of a network in terms of computational complexity, we calculated MAdds. As discussed earlier, the depthwise operation occupies a small portion of the total calculations. Even though ACU is computationally more complex than naive convolution owing to the interpolations, its overhead is not as much as that of the total computations. In addition, the ratio of overheads is reduced with the increasing network width. For example, MobileNet v2($\times$1.4) and ACU($\times$1.2) achieved almost the same accuracies; however, the network with ACU has fewer parameters and complexity. This result shows the possibility that the depthwise ACU can also be used for mobile applications.

\begin{table}[!t]
	\renewcommand{\arraystretch}{1.3}
	\caption{Accuracy of MobileNet v2 with depthwise convolution and depthwise ACU on ImageNet dataset. The accuracy (\%) is evaluated using single-crop images of a single model. The networks using depthwise ACU achieved better accuracy with fewer parameters and complexity. The results are plotted in Fig.~\ref{fig:mobilenet_param_vs_accuracy}. }
	\label{table:result of MobileNet}
	\centering
	\begin{tabular}{c|c|c|c|c}
		\hline
		\textbf{Network}	& \textbf{Top-1}	& \textbf{Top-5} & \textbf{Params} & \textbf{MAdds} \\	\hline \hline			
		MobileNet v2($\times$0.8) & 69.2	& 88.7	& 2.5M	& 201M\\				
		MobileNet v2($\times$1)   & 71.3	& 90.2	& 3.5M	& 314M\\
		MobileNet v2($\times$1.2) & 72.4	& 90.8	& 4.7M	& 437M\\							
		MobileNet v2($\times$1.4) & 73.3	& 91.4	& 6.1M	& 586M\\		\hline
		with depthwise ACU($\times$0.8) & 70.2	& 89.3	& 2.5M	& 244M\\							
		with depthwise ACU($\times$1)   & 72.1	& 90.8	& 3.6M	& 369M\\		
		with depthwise ACU($\times$1.2) & 73.3 & 91.4	& 4.8M	& 502M\\
		with depthwise ACU($\times$1.4) & 74.0 & 91.8	& 6.2M	& 663M\\		\hline		
	\end{tabular}
\end{table}

\subsection{Experiment on CIFAR-10}
We performed an additional experiment on CIFAR-10\cite{krizhevsky2009learning} datasets and compared our method with ResNext~\cite{xie2017aggregated}, which is the representative network using grouped convolution. We initiated with the ResNeXt-29 architecture; This network is 29-layer deep and the width of a bottleneck at the first stage is 512, and the base cardinality is eight. 
We reproduced this network and got slightly less accuracy than the original paper. This is mainly due to the difference in use of the framework: we used \textit{Caffe}\cite{jia2014caffe} but not \textit{Torch}\cite{torch} used in the aforementioned studies. 

From the ResNeXt-29 network, we changed 3$\times$3 grouped convolutions to depthwise ACUs (ACU-29). Even though this change saves lots of parameters (16.13M), the accuracy remains almost the same as that of the base network (Table~\ref{table:result of CIFAR10 resNeXt}). To increase the accuracy with an even deeper network, we stacked five blocks on each stage (ACU-47, 47 layers deep). This network uses 27.37M parameters, which are still less than those in ResNeXt-29, with an error rate of 3.65\%; a similar result was achieved with smaller numbers of parameters.

\begin{table}[!t]
	\renewcommand{\arraystretch}{1.3}
	\caption{Test error of CIFAR-10. Compared to ResNeXt-29, 8$\times$64d model, the use of ACU achieved better result with a smaller number of parameters.}
	\label{table:result of CIFAR10 resNeXt}
	\centering
	\begin{tabular}{c|c|c|c}
		\hline
		\textbf{Network}	& \textbf{Test Error (\%)} 	& \textbf{Params} \\	\hline \hline			
		ResNeXt-29, 8$\times$64d	& 3.95\tablefootnote{Reproduced result. 3.65\% in original paper\cite{xie2017aggregated}. Note that this is the result of final epoch rather than the average of the lowest error they notated.} $\pm$ 0.12	& 34.4M\\	\hline			
		Depthwise ACU-29, $\times$1d & 3.9 $\pm$ 0.09	&16.13M\\	\hline	
		Depthwise ACU-47, $\times$1d & 3.65 $\pm$ 0.06  	& 27.37M\\	\hline	
	\end{tabular}
	
\end{table}

\section{Conclusion}
In this paper, we revisited the active convolution, which we proposed previously~\cite{jeon2017active}, and showed that ACU is an efficient representation of sparse weight convolution. By using an ACU, we can achieve a better result with small number of parameters than that used in naive convolutions. Further experiments have demonstrated that the improvement of accuracy with the use of ACU is due to the ability of training the optimal shape of convolutions.

Beyond ACU, we proposed the grouped ACU, which can use a number of position sets instead of one shared position. By applying groups of positions, multiple receptive fields can be observed in a layer. We also showed that the 1$\times$1 convolution followed by the grouped ACU is a generalization of the Inception module. Furthermore, by increasing cardinalities, we observed that the network with an ACU retains a similar accuracy even while the number of parameters decreases. As a result, we suggest that a depthwise ACU is a good option because it greatly reduces computation cost as well as the number of parameters. Our experiments showed the effectiveness of the depthwise ACU for various benchmarks and networks, and it is considered an attractive unit for replacing naive convolutions.


%

\appendices
\section{Proof of Theorem \ref{thm:equivalent calculation for ACU}} \label{APX:mathmatcial expansion}

In Eq.~\eqref{eq:ACU eq2}, input $x_{c,m+\alpha_k,n+\beta_k}$ can be calculated through bilinear interpolation:

\begin{equation}
\label{eq:interpolated input}
\begin{aligned}
x_{c,m+\alpha_k,n+\beta_k} &= Q^{11}_{c,k}\cdot(1 - \Delta\alpha_k)\cdot(1 - \Delta\beta_k)\\
&+Q^{21}_{c,k}\cdot\Delta\alpha_k\cdot(1- \Delta\beta_k)\\
&+Q^{12}_{c,k}\cdot(1-\Delta\alpha_k)\cdot\Delta\beta_k\\
&+Q^{22}_{c,k}\cdot\Delta\alpha_k\cdot\Delta\beta_k\\
\end{aligned}
\end{equation}
where $Q^{ab}_{c,k}$ represents the four nearest integer points(Fig.~\ref{fig:calc_by_interpolation_extrapolation}):
\begin{equation}
\label{eq:Q def}
\begin{aligned}
Q^{11}_{c,k} = x_{c,m^1_k,n^1_k}, Q^{12}_{c,k} = x_{c,m^1_k,n^2_k},\\
Q^{21}_{c,k} = x_{c,m^2_k,n^1_k}, Q^{22}_{c,k} = x_{c,m^2_k,n^2_k}.
\end{aligned}
\end{equation}
where
\begin{equation}
\label{eq:delta alpha,beta}
\begin{aligned}
\Delta\alpha_k = \alpha_k - \lfloor	{\alpha_k}\rfloor\\
\Delta\beta_k = \beta_k - \lfloor	{\beta_k}\rfloor
\end{aligned}
\end{equation}

\begin{equation}
\label{eq:m1,n1 def_1}
\begin{aligned}
m^1_k = m + \lfloor	{\alpha_k}\rfloor,~& m^2_k = m^1_k + 1,\\
n^1_k = n + \lfloor	{\beta_k}\rfloor,~& n^2_k = n^1_k + 1
\end{aligned}
\end{equation}

\medskip\noindent
By using Eq.~\eqref{eq:interpolated input}, Eq.~\eqref{eq:ACU eq2} is converted to
\begin{equation}
\label{eq:ACU eq2_2}
\begin{aligned}
y_{m,n} = &\sum_{c}\sum_{k} w_{c,k} \cdot \{Q^{11}_{c,k}\cdot(1 - \Delta\alpha_k)\cdot(1 - \Delta\beta_k)\\
&~~~~~~~~~~~~~~~~~~+Q^{21}_{c,k}\cdot\Delta\alpha_k\cdot(1- \Delta\beta_k)\\
&~~~~~~~~~~~~~~~~~~+Q^{12}_{c,k}\cdot(1-\Delta\alpha_k)\cdot\Delta\beta_k\\
&~~~~~~~~~~~~~~~~~~+Q^{22}_{c,k}\cdot\Delta\alpha_k\cdot\Delta\beta_k\}\\
= &\sum_{c}\sum_{k} (1 - \Delta\alpha_k)\cdot(1 - \Delta\beta_k) \cdot w_{c,k} \cdot Q^{11}_{c,k}\\
&~~~~~~~~~~~~~~~~+\Delta\alpha_k\cdot(1- \Delta\beta_k) \cdot w_{c,k} \cdot Q^{21}_{c,k}\\
&~~~~~~~~~~~~~~~~+(1-\Delta\alpha_k)\cdot\Delta\beta_k \cdot w_{c,k} \cdot Q^{12}_{c,k}\\
&~~~~~~~~~~~~~~~~+\Delta\alpha_k\cdot\Delta\beta_k  \cdot w_{c,k} \cdot Q^{22}_{c,k}\\
\end{aligned}
\end{equation}

\medskip\noindent
By substituting Eq.~\eqref{eq:ACU eq2_2} with extrapolated weight $w^{11},w^{21},w^{12}$, and $w^{22}$ :

\begin{equation}
\label{eq:extrapolated weight}
\begin{aligned}
w^{11}_{c,k} = &(1 - \Delta\alpha_k)\cdot(1 - \Delta\beta_k) \cdot w_{c,k}\\
w^{21}_{c,k} = &\Delta\alpha_k\cdot(1- \Delta\beta_k) \cdot w_{c,k}\\
w^{12}_{c,k} = &(1-\Delta\alpha_k)\cdot\Delta\beta_k \cdot w_{c,k}\\
w^{22}_{c,k} = &\Delta\alpha_k\cdot\Delta\beta_k \cdot w_{c,k},
\end{aligned}
\end{equation}

\medskip\noindent
Equation~\eqref{eq:ACU eq2_2} becomes
\begin{equation}
\label{eq:ACU eq2_3}
\begin{aligned}
y_{m,n} = &\sum_{c}\sum_{k} w^{11}_{c,k} \cdot Q^{11}_{c,k}+w^{21}_{c,k} \cdot Q^{21}_{c,k}\\
&~~~~~~~~~+w^{12}_{c,k} \cdot Q^{12}_{c,k}+w^{22}_{c,k} \cdot Q^{22}_{c,k}\\
= & \sum_{c}\sum_{i,j} \overline w_{c,i,j} \cdot x_{c,m+i,n+j},\\
\end{aligned}
\end{equation}
where $|i| \le \max_k(\lceil\alpha_k\rceil), |j| \le \max_k(\lceil\beta_k\rceil)$, and $\overline w_{c,i,j}$ is the sum of extrapolated weights in the overlapping locations, and can be formulated as Eq.~\eqref{eq:summation of extrapolated weight}.

\begin{equation}
\label{eq:summation of extrapolated weight}
\overline w_{c,i,j} = \sum_{k} w_{c,k}^{ab} \cdot I( |i-\alpha_k|<1 \text{~and~} |j-\beta_k| < 1 ),
\end{equation}
where $I(\cdot)$ is an indicator function and

\begin{equation}
\label{eq:summation of extrapolated weight2}
\begin{aligned}
&a =
\begin{cases}
1, & \text{if $\lfloor{\alpha_k}\rfloor = i$} \\
2, & \text{if $\lfloor{\alpha_k}\rfloor+1 = i$} \\
\end{cases}\\
&b =
\begin{cases}
1, & \text{if $\lfloor{\beta_k}\rfloor = j$} \\
2, & \text{if $\lfloor{\beta_k}\rfloor+1 = j$} \\
\end{cases}
\end{aligned}
\end{equation}

\medskip\noindent
Therefore, we can develop extrapolated weight $\boldsymbol{\overline W}_{\theta_p}$, according to which Eq.~\eqref{eq:ACU eq1_3} holds by using $\overline w_{c,i,j}$. 

\begin{equation}
\label{eq:ACU eq1_3}
\begin{aligned}
\boldsymbol{W} * \boldsymbol{X}_{\theta_p} 
= \boldsymbol{\overline W}_{\theta_p} * \boldsymbol{X}
\end{aligned}
\end{equation}
%

\section{Experiment Details} \label{appendix:simple setup}

We trained the network by using 50k images and tested it by using 10k images. The inputs were normalized and randomly cropped to 32$\times$32 images, padded by four pixels on each side and flipped horizontally according to the method in \cite{he2016deep, xie2017aggregated}. The weight parameters were initialized using the method proposed by He et al.~\cite{he2015delving}. The initial learning rate was 0.1 and was divided by 10 after 32k and 48k iterations. The networks were trained with 64k iterations by using stochastic gradient descent with the Nesterov momentum. The batch size was 128, and the momentum was 0.9. We used L2 regularization, and the weight decay was 5e-4. For training ACU, we initialized the positions of the synapses with the shape of a conventional convolution and used a normalized gradient with the learning rate of 1e-3 as position parameter.

The networks consist of three stages, and each stage has three residual blocks (Table \ref{table:simple exp baseline}). After each stage, the widths are doubled, and a 3$\times$3 convolution with stride 2 is applied at the first block of each stage. We used a pre-activation style residual network \cite{he2016identity} with a projection shortcut.

\begin{table}[!h]
	\renewcommand{\arraystretch}{1.3}
	\caption{The base residual network for experiments on CIFAR-10.}
	\label{table:simple exp baseline}
	\centering
	\begin{tabular}{c|c|c}
		\hline
		\textbf{Stage}	& \textbf{Output}	& \textbf{Type} \\	\hline 
		stage0	& 32$\times$32 & 3$\times$3, 64\\	\hline
		stage1	& 32$\times$32 & $\begin{bmatrix}1 \times 1, 32\\3 \times 3, 32\\1 \times 1, 128\end{bmatrix} \times 3$	\\	\hline
		stage2	& 16$\times$16 & $\begin{bmatrix}1 \times 1, 64\\3 \times 3, 64\\1 \times 1, 256\end{bmatrix} \times 3$	\\	\hline
		stage3	& 8$\times$8 & $\begin{bmatrix}1 \times 1, 128\\3 \times 3, 128\\1 \times 1, 512\end{bmatrix} \times 3$	\\	\hline						
		& 1$\times$1 & \begin{tabular}{@{}c@{}}global average-pooling \\ 10-d fc, softmax\end{tabular} \\	\hline
	\end{tabular}
\end{table}

%
%
\ifCLASSOPTIONcompsoc
  \section*{Acknowledgments}
\else
  \section*{Acknowledgment}
\fi
This research was supported in part by National Research Foundation of Korea (NRF) funded by the Korean Government MSIT (NRF-2017R1A2A2A05001400), in part by the Engineering Research Center Program through the National Research Foundation of Korea (NRF) funded by the Korean Government MSIT (NRF-2018R1A5A1059921), and in part by the ICT R\&D program of MSIP/IITP (2016-0-00563, Research on Adaptive Machine Learning Technology Development for Intelligent Autonomous Digital Companion)


%
%
%
%
%



\bibliographystyle{IEEEtran}
\bibliography{egbib}
%

%


\end{document}